\documentclass[sigconf,nonacm,pbalance]{acmart}

\AtBeginDocument{%
  }

\setcopyright{none}
\renewcommand\footnotetextcopyrightpermission[1]{}
\acmConference[MM '26]{Proceedings of the 34th ACM International Conference on Multimedia}{November 10--14, 2026}{Rio de Janeiro, Brazil}

\usepackage{booktabs}
\usepackage{multirow}
\usepackage{graphicx}
\usepackage{xcolor}
\usepackage{tabularx}
\usepackage{cleveref}
\usepackage{placeins}
\usepackage{enumitem}
\usepackage{comment}

\definecolor{goodgreen}{RGB}{0,140,60}

\newlength{\imgw}
\usepackage{array}

\newcommand{\mname}[2]{\textbf{#1}\raisebox{-0.15\height}{\includegraphics[height=1.35em]{#2}}}
\newcommand{\tagt}[1]{\textcolor{black!55}{\texttt{#1}}}
\newcolumntype{L}[1]{>{\raggedright\arraybackslash}p{#1}}
\newcolumntype{Y}{>{\raggedright\arraybackslash}X}
\newcommand{\bad}[1]{{\color{red}#1}}
\newcommand{\good}[1]{{\color{goodgreen}#1}}
\newlength{\imgsep}
\newif\ifincludeappendix
\includeappendixfalse
\begin{document}

\title[GMoT for Micro-Gesture Video Reasoning]{GMoT: Gated Motion-Aware Tokenization for Fine-Grained Micro-Gesture Video Reasoning with Multimodal LLMs}

\author{Taorui Wang}
\orcid{0009-0003-3492-4068}
\affiliation{%
  \institution{Harbin Institute of Technology, Shenzhen}
  \city{Shenzhen}
  \country{China}}
\affiliation{%
  \institution{Great Bay University}
  \city{Dongguan}
  \country{China}}
\email{wangtaorui@stu.hit.edu.cn}

\author{Wei Xia}
\affiliation{%
  \institution{Great Bay University}
  \city{Dongguan}
  \country{China}}

\author{Hui Ma}
\affiliation{%
  \institution{Great Bay University}
  \city{Dongguan}
  \country{China}}

\author{Zijia Song}
\affiliation{%
  \institution{Great Bay University}
  \city{Dongguan}
  \country{China}}

\author{Jiayu Zhang}
\affiliation{%
  \institution{Great Bay University}
  \city{Dongguan}
  \country{China}}

\author{Zeheng Wang}
\affiliation{%
  \institution{Great Bay University}
  \city{Dongguan}
  \country{China}}

\author{Yong Xu}
\authornote{Corresponding authors: Yong Xu and Zitong Yu.}
\affiliation{%
  \institution{Harbin Institute of Technology, Shenzhen}
  \city{Shenzhen}
  \country{China}}
\email{laterfall@hit.edu.cn}

\author{Zitong Yu}
\authornotemark[1]
\affiliation{%
  \institution{Great Bay University}
  \city{Dongguan}
  \country{China}}
\affiliation{%
  \institution{Dongguan Key Laboratory for Intelligence and Information Technology}
  \city{Dongguan}
  \country{China}}
\email{zitong.yu@ieee.org}

\makeatletter
\renewcommand{\@mkauthors}{%
  \begingroup
  \gdef\@currentauthors{}%
  \hsize=\textwidth
  \global\setbox\mktitle@bx=\vbox{%
    \noindent\unvbox\mktitle@bx\par\medskip
    \centering
    {\@authorfont
      \mbox{Taorui Wang\textsuperscript{1,2}}\hspace{1.25em}%
      \mbox{Wei Xia\textsuperscript{2}}\hspace{1.25em}%
      \mbox{Hui Ma\textsuperscript{2}}\hspace{1.25em}%
      \mbox{Zijia Song\textsuperscript{2}}\par
      \smallskip
      \mbox{Jiayu Zhang\textsuperscript{2}}\hspace{1.25em}%
      \mbox{Zeheng Wang\textsuperscript{2}}\hspace{1.25em}%
      \mbox{Yong Xu\textsuperscript{1,*}}\hspace{1.25em}%
      \mbox{Zitong Yu\textsuperscript{2,3,*}}\par}
    \medskip
    {\@affiliationfont
      \textsuperscript{1}Harbin Institute of Technology, Shenzhen, Shenzhen, China\par
      \textsuperscript{2}Great Bay University, Dongguan, China\par
      \textsuperscript{3}Dongguan Key Laboratory for Intelligence and Information Technology, Dongguan, China\par
      \smallskip
      \href{mailto:wangtaorui@stu.hit.edu.cn}{wangtaorui@stu.hit.edu.cn}\quad
      \href{mailto:laterfall@hit.edu.cn}{laterfall@hit.edu.cn}\quad
      \href{mailto:zitong.yu@ieee.org}{zitong.yu@ieee.org}\par}
    \bigskip}%
  \endgroup}
\makeatother

\renewcommand{\shortauthors}{Wang et al.}

\begin{abstract}
Micro-gesture recognition demands the detection of fleeting, spatially localized movements that are frequently overwhelmed by dominant static appearances and background noise. 
While Multimodal Large Language Models (MLLMs) excel at general video understanding, they inherently struggle with subtle kinematics and often rely on static posture priors.
To this end, we propose GMoT, a Gated Motion-Aware Tokenization module that explicitly distills sparse kinematic evidence into a compact sequence prior to temporal modeling. 
GMoT dynamically spotlights action-relevant regions via spatially weighted pooling, extracts adjacent-frame temporal differencing to capture precise motion energy, and adaptively fuses these cues into the visual stream using a conservatively initialized semantic gate.
To transition from simple classification to evidence-grounded reasoning, we further introduce a progressive reward-guided policy refinement paradigm, supported by a semi-supervised annotation pipeline that generates anatomically focused captions.
Beyond achieving the best Top-1 accuracy among the compared methods on iMiGUE (67.32\%) and SMG (73.11\%), improving the Qwen3-VL-8B baseline by +6.80 and +3.11 points, our framework introduces Body-Region Grounding (BRG) Recall as an anatomical-grounding proxy conditioned on correct predictions, together with an overlapping-label cross-domain transfer protocol between iMiGUE and SMG.
Extensive evaluations demonstrate that our GMoT-augmented model improves in-domain accuracy, retains clear gains under label-preserving corruptions, and improves accuracy-oriented cross-domain transfer under explicit small-split caveats while maintaining high anatomical grounding in its generated rationales.
All code will be publicly released in our \href{https://github.com/timwang2001/GMoT-Motion-Aware-Tokenization-for-Fine-Grained-Micro-Gesture-Video-Reasoning-with-Multimodal-LLMs}{GitHub repository}.

\end{abstract}

\begin{teaserfigure}
\centering
\includegraphics[width=0.97\textwidth]{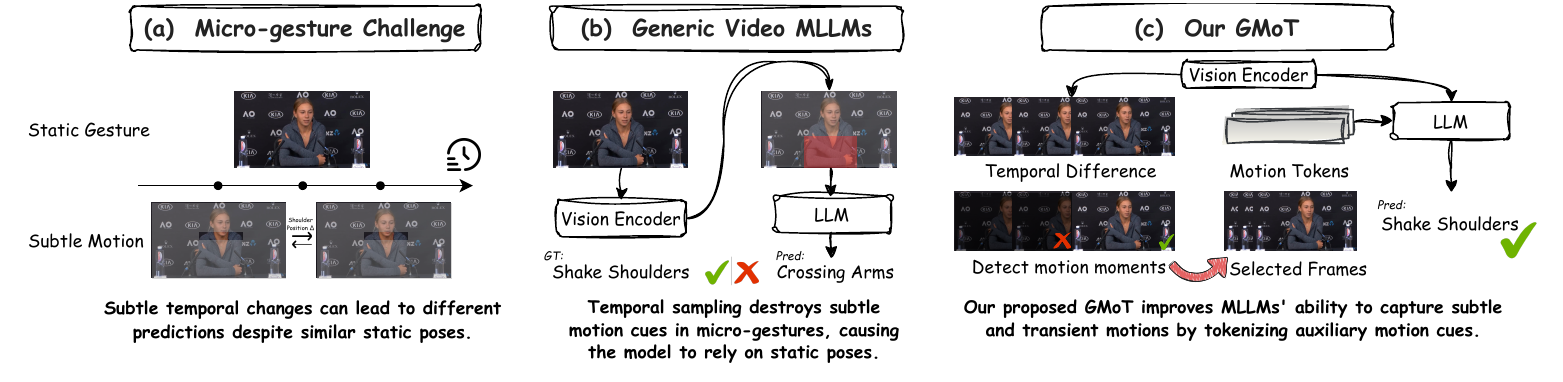}
\caption{\textbf{Problem setting and motivation.} Generic video MLLMs often over-emphasize static appearance, whereas GMoT explicitly distills weak local motion into compact motion-aware tokens for evidence-grounded micro-gesture reasoning.}
\Description{A three-panel teaser figure. Left: micro-gesture clips with subtle, localized short-term motion. Middle: a generic video MLLM focuses on static appearance and makes an appearance-biased mistake. Right: GMoT highlights motion-relevant regions and temporal differences to produce the correct, evidence-grounded prediction.}
\label{fig:teaser}
\end{teaserfigure}

\maketitle

\section{Introduction}
\label{sec:intro}
Micro-gestures (MGs) are subtle, spontaneous, and highly localized human actions that reveal latent affective and cognitive states during daily communication~\cite{axtell1998gestures,liu_imigue_2021,chen2019analyze,wang2026mgsurvey}. 
Unlike macroscopic actions (e.g., running or jumping), MGs, such as a brief finger twitch, a slight shoulder shrug, or a tense lip press, manifest in extremely narrow spatial regions and occur across just a few frames~\cite{chen_smg_2023,wang2026mgsurvey}. 
Accurately recognizing these movements provides fine-grained behavioral cues essential for downstream applications such as emotion understanding, psychological interaction analysis, and multimodal deception detection~\cite{li_msf-mamba_2025,patapati2025clipmgguidingsemanticattention,li_deemo_2025,zhang2026multimodal}.
However, visually similar categories often differ only in subtle kinematic details, which makes reliable discrimination under real-world conditions particularly challenging.

Recently, Multimodal Large Language Models (MLLMs) have revolutionized video understanding by coupling powerful visual encoders with the reasoning capabilities of Large Language Models (LLMs)~\cite{liu2023visualinstructiontuning,bai2023qwenvlversatilevisionlanguagemodel}. 
In theory, MLLMs offer an attractive foundation for micro-gesture recognition, which can go beyond merely outputting generic class labels to connect visual evidence with language-based explanations and articulate what moved and how it changed~\cite{li_deemo_2025,li2025ealdmllmemotionanalysislongsequential}.
However, directly applying existing MLLMs to micro-gesture recognition often yields suboptimal and misleading results, because their reliance on dense frame-level spatial patch extraction causes weak and sparse kinematic signals to be either missed by sparse sampling or diluted by temporal pooling~\cite{chen2023videollmmodelingvideosequence,bai2023qwenvlversatilevisionlanguagemodel}.
Without sufficient temporal sensitivity, models tend to resort to what we term \textit{hasty guessing}, namely making quick but unreliable judgments based on salient yet coarse appearance patterns before capturing fine-grained motion evidence, for example, classifying a crossed-arm posture while overlooking a subtle but discriminative finger tap.
Even worse, because standard paradigms only reward the final classification accuracy, as~\Cref{fig:teaser} shows, the models are incentivized to hallucinate plausible-sounding but visually ungrounded reasoning traces.
Moreover, most existing evaluations emphasize within-dataset Top-1 accuracy, leaving it unclear whether a model truly captures transferable micro-dynamics or merely memorizes dataset-specific appearance biases.

These critical limitations highlight an urgent need for a paradigm shift from purely label-driven recognition to evidence-based kinematic reasoning. 
To achieve this, the MLLM requires a strong inductive bias that explicitly forces the model to focus on localized temporal changes rather than static visual contexts.
To this end, we propose \textbf{GMoT}, a lightweight and highly targeted \textbf{G}ated \textbf{M}otion-Aware \textbf{T}okenization module that distills sparse spatio-temporal cues into a compact motion-sensitive token sequence without relying on computationally heavy 3D convolutions or dense space-time attention that easily overfits to spatial biases.
Specifically, GMoT establishes a spatial saliency scorer to spotlight action-relevant regions dynamically, computes explicit adjacent-frame temporal differences to isolate precise kinematic energy, and injects these cues into the MLLM's visual stream via a conservatively initialized adaptive semantic gate.
Furthermore, we argue that evaluating micro-gesture recognition requires moving beyond Top-1 accuracy within a single dataset.
To improve the visual grounding of generated explanations, we build a semi-supervised annotation pipeline to generate Chain-of-Thought (CoT) descriptions focused on observable motion evidence.
We then optimize the MLLM using a progressive reward-guided policy refinement strategy~\cite{shao2024deepseekmathpushinglimitsmathematical}, guided by a reward formulation that penalizes lazy guessing and encourages anatomical region focus.
To quantify one transparent aspect of these capabilities, we introduce Body-Region Grounding (BRG) Recall, a lexical anatomical-grounding proxy computed on correctly classified samples. We further establish an overlapping-label cross-domain transfer protocol between iMiGUE and SMG as a diagnostic of robustness under dataset shift.

Our contributions are summarized as follows:
\begin{itemize}[leftmargin=2em, topsep=4pt, itemsep=4pt, parsep=0pt]
    \item \textbf{Motion-aware tokenization.} We propose GMoT, which injects motion-salient pooling, adjacent-frame differencing, and gated fusion into an MLLM without disrupting its pretrained visual-language interface.
    \item \textbf{Reasoning-oriented training.} We construct semi-automatic CoT supervision and a progressive reward-guided refinement recipe that encourages label correctness, grounded explanations, and resistance to lazy appearance-based guessing.
    \item \textbf{Broader evaluation.} We introduce BRG Recall as a scoped anatomical-grounding proxy and an overlapping-label cross-domain transfer protocol between iMiGUE and SMG. GMoT reaches 67.32\% / 73.11\% Top-1 accuracy, respectively, and remains stronger than its Qwen3-VL backbone under the tested label-preserving corruptions.
\end{itemize}

\section{Related work}

\subsection{Micro-gesture Recognition}

Recent progress in micro-gesture recognition (MGR) has been largely shaped by emerging datasets and the evolution of modeling choices across modalities and learning paradigms~\cite{li_deemo_2025,li_msf-mamba_2025,wang2026affectagentcollaborativemultiagentreasoning}. RGB-based approaches typically adapt spatio-temporal backbones to capture subtle motion cues, while skeleton-based methods model joint dynamics with sequential models or graph networks~\cite{li_msf-mamba_2025,li2023jointskeletalsemanticembedding}. To handle highly similar classes and noisy observations, later work increasingly emphasizes stronger temporal attention and multi-modal fusion between appearance and pose signals~\cite{patapati2025clipmgguidingsemanticattention,huang2024multi,gu2025mm,chen2024prototype}. Despite these advances, most studies still cast the task as closed-set classification and therefore provide limited support for evidence-grounded micro-gesture reasoning.
Broader surveys of generative affective modeling and human motion video generation further highlight the growing role of generative and language models in human-centered video analysis~\cite{ma2025generative,xue2025humanmotion,wang2026navigatingemotiontreehierarchical,wang2026dfm}.

\subsection{MLLMs for Video Reasoning}
MLLMs have recently extended from image understanding to video by combining vision encoders with language decoders and training on instruction-following data~\cite{bai2023qwenvlversatilevisionlanguagemodel,li2024mvbench,zhu2025emosym,zhu2025uniemo,ye2026eyes}. 
Video MLLMs combine vision encoders with language decoders under strict token budgets, so they often rely on sparse sampling, temporal pooling, or lightweight temporal adapters~\cite{bai2023qwenvlversatilevisionlanguagemodel,li2024mvbench,chen2025lvagent,zeng2024timesuite,zhu2026delta,zhu2026H-GAR,zhang2026mavis,wang2026rsiccllm}. These designs are effective for coarse video understanding, yet they often blur the brief localized dynamics that distinguish MG classes. Domain-specific systems have begun to explore LLMs for subtle behavioral understanding: AU-LLM maps fused 3D-CNN features to a compact visual token for micro-expression action-unit detection, while DeceptionX organizes fine-grained audiovisual evidence into interpretable deception reasoning~\cite{liu2025llm,zhang2026deceptionx}. Recent systems further extend LLM-based multimodal learning to cross-modal physiological sensing and rationale-driven affective reasoning~\cite{xie2026physllm,cheng2026omniopsd}, while Reflect-R1 grounds self-correction in retrieved visual evidence for long-video understanding~\cite{chen2026reflectr1}. Complementary continual-learning methods use predictive prompting and task-oriented generation to balance adaptation and retention across class increments~\cite{huang2026preprompt,huang2024etag}. Although GMoT is not designed for class-incremental learning, these methods provide a broader view of controlled prompt- and representation-level adaptation. Nevertheless, most video MLLMs
are optimized for answer correctness rather than evidence tracing,
making it hard to produce reliable and interpretable decisions for
subtle micro-gestures.

\section{Methodology}
In this section, we first introduce the preliminaries of multimodal large language models. We then present \textbf{GMoT}, a gated motion-aware token module that enables a vanilla MLLM to capture subtle temporal cues from micro-gesture videos by distilling frame-level visual tokens into compact motion representations. Finally, we describe a four-stage training strategy that progressively aligns classification, shapes the chain-of-thought pattern, and enhances micro-gesture reasoning via reward-guided policy refinement.

\begin{figure*}[ht]
  \centering
  \includegraphics[width=0.93\linewidth]{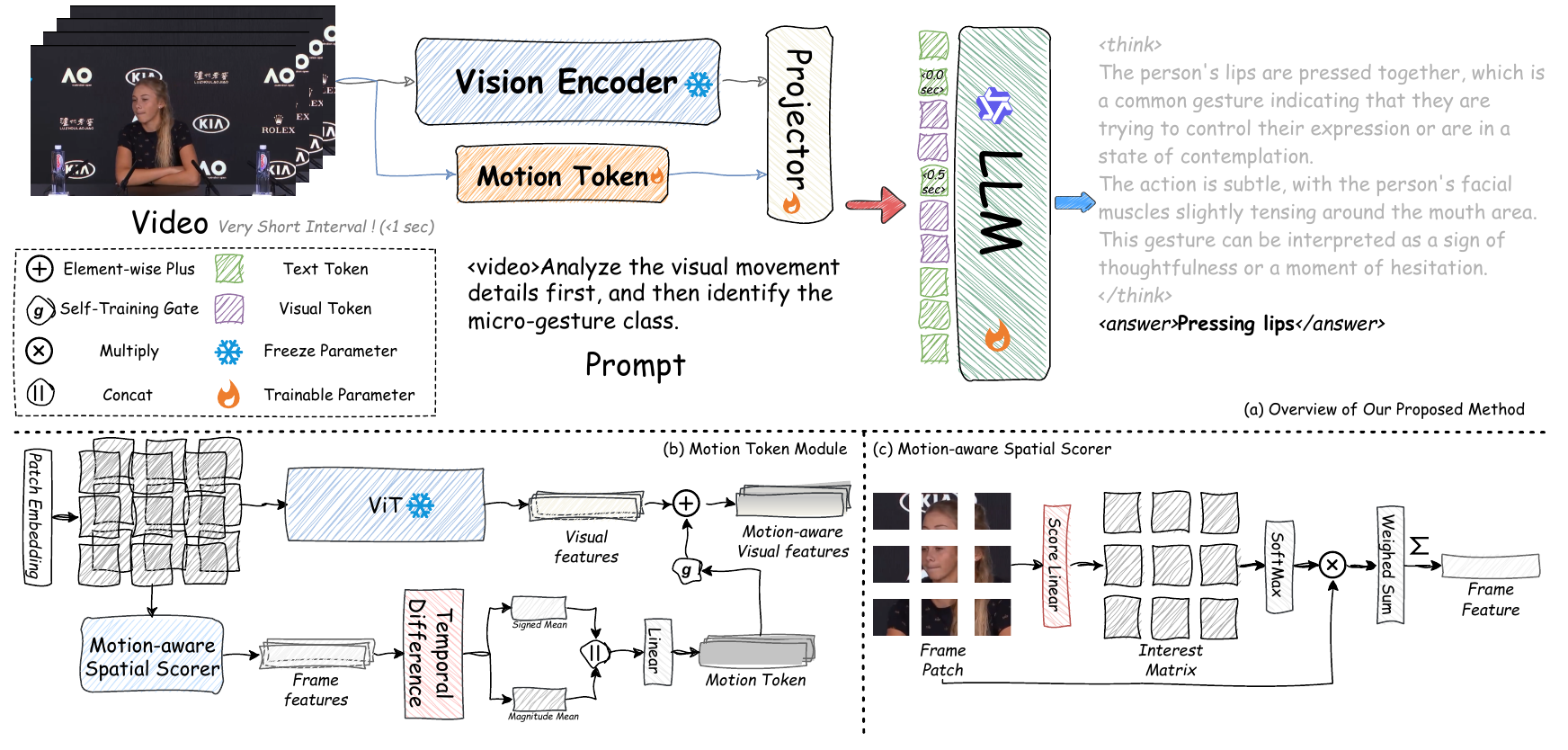}
  \caption{\textbf{Overview of GMoT.} GMoT extracts motion-sensitive evidence from short video clips and injects the resulting motion token into the MLLM through a gated fusion pathway for motion-aware reasoning. (a) presents the overall pipeline, where a compact motion token is derived from a short clip and incorporated into the MLLM together with visual and textual inputs. (b) illustrates the motion token module, which integrates motion-guided spatial saliency, adjacent-frame temporal differencing, and a learnable self-training gate. (c) shows the motion-aware spatial scorer, which predicts a soft importance map over frame patches and aggregates them into frame-level features.}

  \Description{Method overview showing spatial scoring, adjacent-frame differencing, motion token generation, and gated fusion into the visual stream.}
  \label{fig:methodology}
\end{figure*}

\subsection{Preliminary}
\textbf{Multimodal large language models.}
Since the introduction of LLaVa~\cite{liu2023visualinstructiontuning}, multimodal large language models (MLLMs) have attracted increasing attention from the community. Among recent advances, Qwen3-VL~\cite{bai2025qwen3vltechnicalreport} has become a strong and widely adopted backbone because of its strong multimodal understanding capability. Following the LLaVa paradigm, the dense version of Qwen3-VL has three modules: a SigLIP-2 architecture vision encoder~\cite{tschannen2025siglip2multilingualvisionlanguage}, an MLP-based aligner, and the Qwen3 LLM~\cite{yang2025qwen3technicalreport}.

To process an image or a video, the vision encoder first converts each frame into a sequence of patch-level visual tokens. Given a video clip $V=\{I_t\}_{t=1}^{T}$, the encoder produces patch tokens $\mathbf{X}_t \in \mathbb{R}^{N \times d_v}$ for each frame, and the aligner maps them into the language space as $\mathbf{Z}_t \in \mathbb{R}^{N \times d}$.

With the LLaVA-style multimodal prompting, the aligned visual tokens are concatenated with textual tokens and fed into the LLM:
\begin{equation}
\mathbf{H} = \mathrm{LLM}\big([\mathbf{Z}_1,\ldots,\mathbf{Z}_T;\ \mathbf{E}(\mathbf{p})]\big),
\end{equation}
where $\mathbf{p}$ is the input prompt and $\mathbf{E}(\cdot)$ denotes the text tokenizer. The model finally generates an output sequence $\hat{\mathbf{y}}$ in an autoregressive manner, which can be either a class name in a recognition task or a reasoning-augmented response in a reasoning scenario.

However, directly encoding patch tokens from frames may dilute subtle micro-gesture signals. The temporal variation of micro-gestures often exhibits low amplitude and short duration, which are easily overlooked during temporal modeling or diluted during visual token aggregation. This motivates us to distill time-sensitive information into an auxiliary representation that can be seamlessly injected into the MLLM. 

\subsection{Semi-automatic Reasoning Annotation}
We construct reasoning-aware supervision with the semi-automatic pipeline illustrated in the left part of~\Cref{fig:training_strategy}. For each labeled clip, two independent LMMs generate candidate motion descriptions under prompts that suppress background and identity cues while emphasizing action-relevant body regions. Qwen3-VL-MoE~\cite{bai2025qwen3vltechnicalreport} then selects the better candidate using class correctness, visual grounding, and non-hallucination, conditioned on a class-level reference description. Borderline cases are manually reviewed, and a human validation study based on sampled accepted descriptions is summarized in Appendix~A. The resulting annotations supervise both label prediction and evidence-grounded CoT generation.

More concretely, for each clip, using two candidates improves diversity and reduces the risk of under-specified or hallucinated details from a single case. In prompts, we explicitly discourage any description of background, scene context, or identity-related facial information. Instead, the models are required to focus on action-relevant regions, with primary attention to the hands and arms, and to describe only observable motion patterns that support the MG label. We additionally provide a class-level description from Deepseek~\cite{shao2024deepseekmathpushinglimitsmathematical} as a reference, summarizing typical visual patterns and key regions for each MG category. This serves as a semantic anchor to check class consistency and reduce label drift.

\subsection{GMoT: Gated Motion-Aware Tokenization Module}

While vanilla MLLMs are strong at general video understanding, they still struggle with fleeting and localized MG cues. As illustrated in~\Cref{fig:methodology}, \textbf{GMoT} addresses this mismatch by injecting explicit kinematic bias into the visual stream and distilling sparse spatio-temporal evidence into compact motion-sensitive tokens.

\subsubsection{Motion-Prompted Spatial Saliency.} Given the localized nature of micro-gestures, spatially homogeneous pooling will dilute the discriminative cues with irrelevant background noise. Instead of relying on the LLM to implicitly search for these tiny regions across dense patches, GMoT establishes a lightweight spatial scorer to explicitly spotlight motion-relevant areas. Specifically, we project each patch token from the vision encoder through a linear layer to obtain a relevance score. A spatial softmax is then applied to produce normalized attention weights, dynamically aggregating patch features into a frame-level representation via weighted summation. This mechanism functions as a learned spatial filter, aggressively amplifying localized evidence while suppressing visual redundancy.

\subsubsection{Kinematics-Inspired Temporal Differencing.} To capture the precise moment a micro-gesture occurs, the model must be sensitive to temporal gradients rather than static appearances. While standard video transformers rely on complex temporal mixing, we argue that explicit temporal differencing offers a more direct and robust inductive bias for subtle motion. 
Rather than treating this difference as a monolithic feature, we decouple it into a two-stream kinematic representation. We compute the mean magnitude to explicitly quantify motion energy, while retaining the signed difference to preserve the directional kinematics. By concatenating these decoupled streams and applying a learnable projection, we compress the raw temporal gradients into a fixed-length, dense motion token. This design forces the network to focus exclusively on visual changes, effectively isolating micro-dynamics from static human appearances.
\subsubsection{Semantic-Preserving Adaptive Gating.} A critical challenge in adapting pretrained MLLMs is that abruptly injecting new modality tokens can disrupt the aligned visual-semantic space. We therefore use a near-closed gated residual as a stabilization component rather than treating the gate itself as a standalone architectural novelty. Let $z_t$ denote the aligned original visual feature at time $t$, $g$ denote the learned gate parameter, and $m_t$ denote the distilled motion token. We fuse them through
\begin{equation}
\tilde{\mathbf{z}}_t = \mathbf{z}_t + \sigma(\mathbf{g})\odot \mathbf{m}_t,
\end{equation}
where $\sigma$ is the sigmoid function and $\odot$ denotes element-wise multiplication. The gate is initialized in its near-closed regime so that $\tilde{z}_t \approx z_t$ at the start of adaptation. This conservative starting point preserves the pretrained representation while allowing the model to learn how much motion evidence to introduce. The contribution of GMoT lies in the task-specific composition of localized spatial scoring, explicit temporal differencing, and conservative residual injection before language reasoning.

\begin{figure*}[ht]
  \centering
  \includegraphics[width=0.78\linewidth]{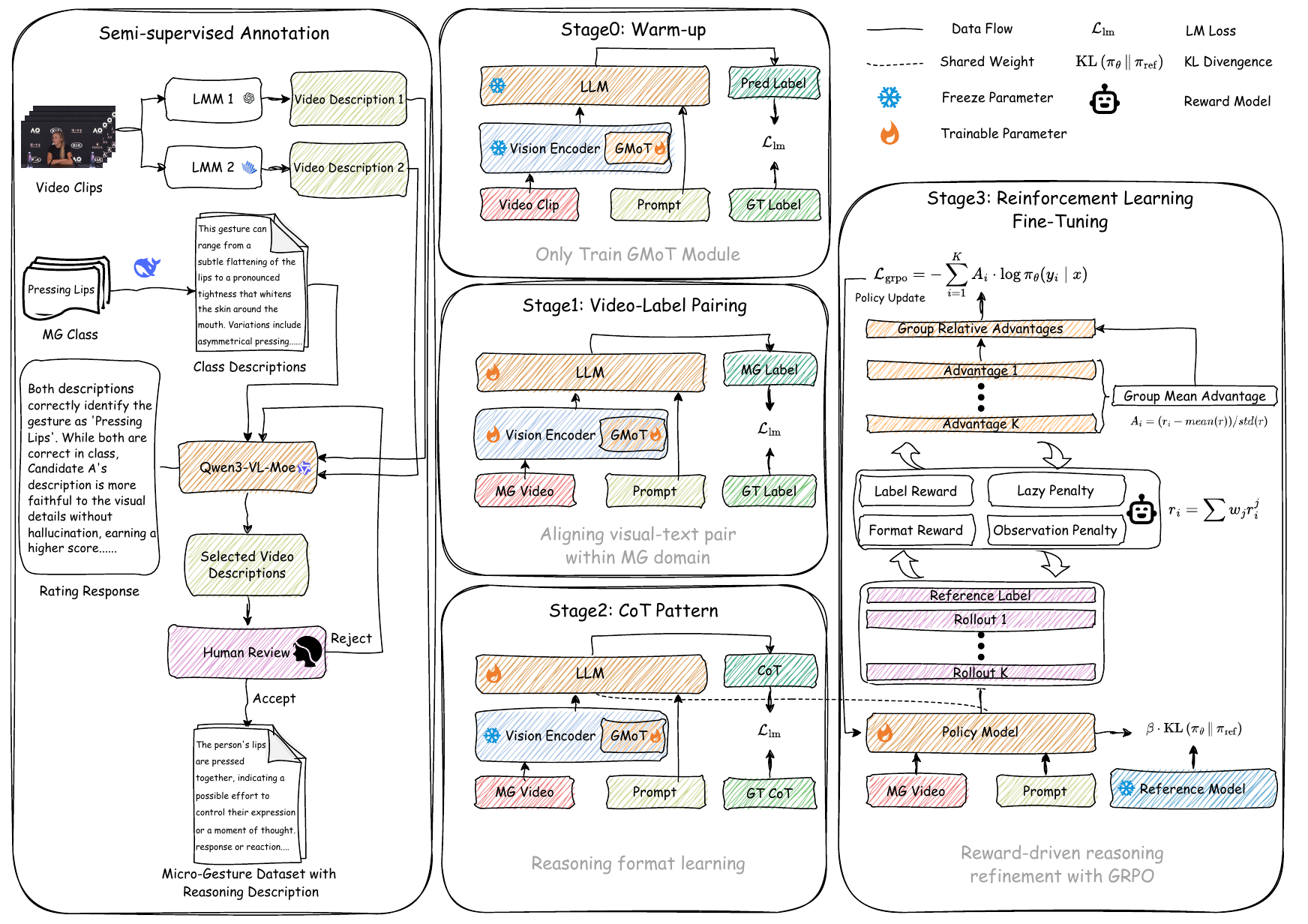}
  \caption{\textbf{Overview of the four-stage training schedule.} The training process progressively moves through GMoT warm-up, micro-gesture domain adaptation, CoT-style reasoning learning, and finally reward-guided reinforcement fine-tuning for reasoning refinement.}
  \Description{Training schedule diagram with four stages from module warm-up to reward-guided policy refinement.}
  \label{fig:training_strategy}
\end{figure*}
\subsection{Training Strategy}
Building on the annotation pipeline and GMoT, we adopt a staged recipe that separates motion token learning, domain adaptation, reasoning format acquisition, and policy refinement, as shown in ~\Cref{fig:training_strategy}. This decomposition avoids destructive interference, where sparse label-only supervision proves insufficient for newly introduced motion tokens. Specifically, we adopt a four-stage schedule.

\subsubsection{Stage 0: GMoT Warm-up.}
We initialize from a pretrained MLLM and warm up the proposed GMoT module. During this stage, we freeze the LLM and the vision encoder and update only the GMoT module on general video datasets. Given a video clip and a prompt, the model is trained to predict the ground-truth label tokens under the standard language modeling loss,
\begin{equation}
\mathcal{L}_{\mathrm{lm}} = - \sum_{t} \log \pi_{\theta}(s_t \mid x, s_{<t}),
\end{equation}
which encourages GMoT to produce motion-aware tokens while keeping the pretrained model stable.

\subsubsection{Stage 1: Video-Label Pairing.}
\label{stage1}
We then adapt the model to the micro-gesture domain with supervised fine-tuning on MG video and label pairs. In this stage, we allow the MLLM components to be trainable in the LoRA~\cite{hu2021loralowrankadaptationlarge} setting and continue to optimize the language modeling objective to align visual representations with MG-specific label semantics. This stage primarily improves domain adaptation and reduces the gap between generic video understanding and micro-gesture recognition.

\subsubsection{Stage 2: CoT-format SFT}
\label{stage2}
Next, we teach the model a consistent reasoning pattern using CoT-style annotations. We freeze the vision encoder to avoid drifting visual features and fine-tune the LLM together with GMoT to generate the target CoT text under the language modeling loss. The supervision is constructed to focus on observable motion evidence rather than background or identity cues, so the model learns an evidence-grounded reasoning path that will be further reinforced in Stage 3.

\subsubsection{Stage 3: Reward-guided Policy Refinement.}
\label{stage3}
After Stage 2, the model already follows a structured reasoning format, but SFT still optimizes token-level likelihood rather than directly favoring globally better responses. We therefore perform a reward-guided policy refinement stage with grouped rollouts and a reference-regularized update objective~\cite{shao2024deepseekmathpushinglimitsmathematical,feng2025videor1reinforcingvideoreasoning} to improve reasoning reliability and reduce failure cases on subtle micro-gestures.

\subsubsection{Reward Formulation.}
Our Stage 3 reward combines label correctness, lazy-prediction penalty, format compliance, and observation-focus terms:
\begin{equation}
    R_i = R^{\mathrm{label}}_i + R^{\mathrm{lazy}}_i + R^{\mathrm{format}}_i + R^{\mathrm{focus}}_i
\end{equation}
\noindent\textbf{Accuracy Reward ($R^{\mathrm{label}}$).}
To prevent the MLLM from hallucinating non-existent micro-gesture concepts, we strictly constrain the output space. Let $\mathcal{C}$ denote the set of standard categories, $c$ be the ground-truth class, and $\hat{c}_i$ be the predicted class parsed from the \texttt{<answer>} field. The label reward applies a large positive signal for exact matches, a standard penalty for incorrect guesses, and a severe penalty for out-of-domain hallucinations:
\begin{equation}
R^{\mathrm{label}}_i = \begin{cases}
+3.0, & \hat{c}_i = c \\
-1.0, & \hat{c}_i \neq c \land \hat{c}_i \in \mathcal{C} \\
-2.0, & \hat{c}_i \notin \mathcal{C}
\end{cases}
\end{equation}

\noindent\textbf{Lazy-Prediction Penalty ($R^{\mathrm{lazy}}$).}
During early training stages, the model tends to exploit the long-tailed distribution by repeatedly guessing high-frequency head classes such as \textit{Illustrative hand gestures}. Let $\mathcal{H} \subset \mathcal{C}$ denote the predefined high-frequency set of such classes. We penalize incorrect predictions that fall into this set:
\begin{equation}
R^{\mathrm{lazy}}_i = \begin{cases}
-0.5, & \hat{c}_i \neq c \land \hat{c}_i \in \mathcal{H} \\
0, & \text{otherwise}
\end{cases}
\end{equation}

\noindent\textbf{Formatting Reward ($R^{\mathrm{format}}$).}
We encourage the completion to follow the CoT template:
\begin{equation}
R^{\mathrm{format}}_i = \delta_{\mathrm{think}} + \delta_{\mathrm{answer}}
\end{equation}
where $\delta_{\mathrm{think}} = 0.5$ if the \texttt{<think>} block is present and $-0.5$ otherwise, while $\delta_{\mathrm{answer}} = 0$ if the \texttt{<answer>} block is well formed and $-1.5$ if it is missing.

\noindent\textbf{Observation Penalty ($R^{\mathrm{focus}}$).}
We penalize background-oriented hallucination in the reasoning trace. If the \texttt{<think>} field contains irrelevant environmental keywords $\mathcal{K}_{\mathrm{bg}}$, then
\begin{equation}
R^{\mathrm{focus}}_i = \begin{cases}
-1.0, & \exists\, k \in \mathcal{K}_{\mathrm{bg}} \text{ appearing in } \texttt{<think>} \\
0, & \text{otherwise}
\end{cases}
\end{equation}

\noindent\textbf{Group rollouts and loss.}
For each input $x$, we sample $K$ completions, compute group-normalized advantages
\begin{equation}
A_i=\frac{r_i-\mu}{\sigma+\epsilon},\quad
\mu=\frac{1}{K}\sum_{i=1}^{K}r_i,\quad
\sigma=\sqrt{\frac{1}{K}\sum_{i=1}^{K}(r_i-\mu)^2},
\end{equation}
and optimize
\begin{equation}
\mathcal{L}_{\mathrm{stage3}}
= -\frac{1}{K}\sum_{i=1}^{K} A_i \log \pi_\theta(y_i \mid x)
+ \beta\, D_{\mathrm{KL}}\!\left(\pi_\theta(\cdot\mid x)\,\|\,\pi_{\mathrm{ref}}(\cdot\mid x)\right),
\label{eq:stage3_objective}
\end{equation}
where the reference policy $\pi_{\mathrm{ref}}$ is adapted from Stage 2. Appendix~B provides the full keyword sets and auxiliary details.

Each reward term targets a specific failure mode that we repeatedly observed in MG reasoning. $R^{\mathrm{label}}$ keeps the model inside the valid label space, $R^{\mathrm{lazy}}$ focuses on long-tail classes, $R^{\mathrm{format}}$ stabilizes the RL template, and $R^{\mathrm{focus}}$ suppresses background-oriented hallucinations. Together with group-normalized advantages, Stage 3 optimizes relative response quality within each prompt rather than raw token regression, while the KL term preserves the grounded reasoning behavior already established in Stage 2.


\section{Experiments}
\subsection{Implementation Details}
We build our method on Qwen3-VL~\cite{bai2025qwen3vltechnicalreport} and all stages are implemented with \texttt{ms-swift}~\cite{zhao2025swiftascalablelightweightinfrastructure} and DeepSpeed ZeRO-3~\cite{rajbhandari2020zeromemoryoptimizationstraining} on NVIDIA H100 GPUs. Since most video clips in the datasets are shorter than 3 seconds, we sample each clip at 4 FPS with at most 12 frames, cap the pixels per frame at 786,432, and limit visual tokens to 896. Appendix~B provides the complete training details.

\begin{table}[t]
    \centering
    \footnotesize
    \setlength\tabcolsep{3.5pt}
    \caption{Top-1 accuracy on iMiGUE~\cite{liu_imigue_2021} and SMG~\cite{chen_smg_2023}.}
    \begin{tabular}{lccc}
    \toprule
    \multirow{2}{*}{\textbf{Method}} & \multirow{2}{*}{\textbf{Model Type}} & \multicolumn{1}{c}{\textbf{iMiGUE}} & \multicolumn{1}{c}{\textbf{SMG}} \\
    & & \textbf{Top-1 Acc. (\%)$\uparrow$} & \textbf{Top-1 Acc. (\%)$\uparrow$} \\
    \midrule
    C3D~\cite{tran2015learning} & CNN & 30.13 & 36.07 \\
    I3D~\cite{carreira2017quo}  & CNN & 35.08 & 28.36 \\
    TSN~\cite{wang2018temporal} & CNN & 51.31 & 50.49 \\
    TSM~\cite{lin2019tsm}       & CNN & 56.23 & 58.69 \\
    MA-Net~\cite{guo2024benchmarking} & CNN & 58.08 & 48.69 \\
    \midrule
    P\&C \cite{su2020predict}          & Encoder-Decoder & 31.67 & 39.30 \\
    U-S-VAE \cite{liu_imigue_2021}       & Encoder-Decoder & 32.43 & 39.59 \\
    \midrule
    Timesformer~\cite{bertasius2021spacetimeattentionneedvideo} & Transformer & 50.66 & 47.89 \\
    Video SwinT~\cite{liu2021videoswintransformer}             & Transformer & 60.11 & 48.03 \\
    UniformerV2~\cite{li2023uniformerv2}  & Transformer & 59.91 & 46.07 \\
    \midrule
    VideoMamba~\cite{li2024videomamba} & SSM & 58.13 & 53.28 \\
    MSF-Mamba-T$^{+}$~\cite{li_msf-mamba_2025}& SSM & 58.33 & 52.29 \\
    MSF-Mamba-M$^{+}$~\cite{li_msf-mamba_2025}& SSM & 62.98 & 60.13 \\
    \midrule
    GPT-4o~\cite{hurst2024gpt} & MLLM & 56.89 & 60.71 \\
    Gemini 2.5 Pro~\cite{comanici2025gemini25} & MLLM & 58.41 & 62.57 \\
    InternVL2.5-8B~\cite{chen2024internvl25} & MLLM & 59.25 & 56.21 \\
    Video-LLaMA2~\cite{damonlpsg2023videollama} & MLLM & 39.74 & 51.80 \\
    TinyLLaVA-Video~\cite{zhang2025tinyllavavideosmallerlmmsvideo} & MLLM & 57.82 & 52.87 \\
    Affect-GPT~\cite{lian2025affectgpt} & MLLM & 58.45 & 54.12 \\
    Qwen2.5-VL-7B~\cite{bai2025qwen25vltechnicalreport} & MLLM & 59.71 & 53.27\\
    Qwen3-VL-2B~\cite{bai2025qwen3vltechnicalreport} & MLLM & 51.25 & 20.98 \\
    Qwen3-VL-8B~\cite{bai2025qwen3vltechnicalreport} & MLLM & 60.52 & 70.00 \\
    \midrule
    Qwen2.5-VL-7B+\textbf{GMoT} & MLLM & 63.23 & 60.85 \\
    Qwen3-VL-2B+\textbf{GMoT} & MLLM & 58.51 & 68.85 \\
    Qwen3-VL-8B+\textbf{GMoT} & MLLM & \textbf{67.32} & \textbf{73.11} \\
    \bottomrule
    \end{tabular}
    \label{tab:Compara}
    
\end{table}

\subsection{Quantitative Results}
\noindent\textbf{Intra-dataset testing.} \Cref{tab:Compara} compares our \textbf{GMoT}-enhanced MLLM with representative micro-gesture recognition approaches on iMiGUE and SMG datasets. All trainable open-source MLLM baselines use the same iMiGUE and SMG training splits. CNN-based backbones such as C3D~\cite{tran2015learning} and I3D~\cite{carreira2017quo} perform poorly, suggesting that purely local spatio-temporal convolutions are insufficient to capture subtle and instantaneous cues. Stronger CNN variants like TSN~\cite{wang2018temporal}, TSM~\cite{lin2019tsm}, and MA-Net~\cite{guo2024benchmarking} improve steadily, yet still lag behind stronger temporal modeling on fine-grained MG settings. Transformer-based models benefit from global attention, but they remain limited on SMG, where weak motion evidence is easily overwhelmed by irrelevant appearance variation. Unsupervised learning methods, e.g., P\&C~\cite{su2020predict} and U-S-VAE~\cite{liu_imigue_2021}, benefit from label-free training but typically achieve suboptimal accuracy.

State-space models offer a stronger trade-off between accuracy and efficiency, with the VideoMamba~\cite{li2024videomamba} and MSF-Mamba~\cite{li_msf-mamba_2025} variants consistently outperforming most CNN and Transformer baselines. On top of this landscape, MLLMs provide a competitive alternative by combining strong visual encoders with language-centric reasoning priors. With significant pretraining data, Qwen2.5-VL-7B~\cite{bai2025qwen25vltechnicalreport} reaches 59.71\% on iMiGUE and 53.27\% on SMG, while Qwen3-VL-8B~\cite{bai2025qwen3vltechnicalreport} further improves to 60.52\% and 70.00\%. Gemini 2.5 Pro~\cite{comanici2025gemini25} and InternVL2.5-8B~\cite{chen2024internvl25} provide additional closed- and open-model reference points. GMoT improves Qwen2.5-VL-7B to 63.23\% and 60.85\%, and Qwen3-VL-8B to 67.32\% and 73.11\%, showing positive portability across backbones and model scales. Appendix~C provides the extended per-family discussion.

For the transfer and reasoning analyses below, we focus on reasoning-capable MLLM backbones evaluated under a unified prompting interface because matched transfer settings and rationale outputs are not currently available for the traditional non-MLLM baselines in \Cref{tab:Compara}. 

\begin{table*}[ht]
\centering
\small
\setlength{\tabcolsep}{4.5pt}
\caption{Cross-domain transfer results on iMiGUE~\cite{liu_imigue_2021} and SMG~\cite{chen_smg_2023}.}
\label{tab:crossdomain_results}
\begin{tabular}{lcccccc}
\toprule
& \multicolumn{3}{c}{iMiGUE $\rightarrow$ SMG ($N=22$)} & \multicolumn{3}{c}{SMG $\rightarrow$ iMiGUE ($N=312$)} \\
\cmidrule(lr){2-4}\cmidrule(lr){5-7}
Model & Acc. (\%) & Macro-F1 & Weighted-F1 & Acc. (\%) & Macro-F1 & Weighted-F1 \\
\midrule
Affect-GPT~\cite{lian2025affectgpt} & 4.55 & 0.0217 & 0.0040 & 5.13 & 0.0195 & 0.0050 \\
Video-LLaMA2~\cite{damonlpsg2023videollama} & 4.55 & 0.0235 & 0.0053 & 0.00 & 0.0000 & 0.0000 \\
MM-Gesture RGB~\cite{gu2025mm} & 13.64 & 0.0952 & 0.1558 & 21.79 & 0.1426 & 0.2217 \\
Qwen2.5-VL-7B~\cite{bai2025qwen25vltechnicalreport} & 50.00 & 0.3743 & 0.5578 & 50.00 & 0.2894 & 0.5034 \\
Qwen3-VL-8B~\cite{bai2025qwen3vltechnicalreport} & 81.82 & 0.3325 & \textbf{0.8130} & 61.86 & 0.3335 & \textbf{0.6762} \\

\midrule
Qwen2.5-VL-7B+\textbf{GMoT} & 77.27 & 0.3457 & 0.7786 & 52.24 & 0.2720 & 0.5378 \\
Qwen3-VL-8B+\textbf{GMoT} & \textbf{86.36} & \textbf{0.4257} & 0.8053 & \textbf{62.18} & \textbf{0.3371} & 0.6720 \\
\bottomrule
\end{tabular}
\end{table*}

\noindent\textbf{Cross-domain transfer.}
\Cref{tab:crossdomain_results} reports the transfer results in two directions. Qwen3-VL-8B+\textbf{GMoT} achieves the best Accuracy and Macro-F1 in both directions. The metric ranking is nevertheless mixed: vanilla Qwen3-VL-8B retains a slightly higher Weighted-F1 on iMiGUE$\rightarrow$SMG (0.8130 vs.\ 0.8053) and SMG$\rightarrow$iMiGUE (0.6762 vs.\ 0.6720). These decreases are 0.0077 and 0.0042, respectively, while Accuracy and Macro-F1 improve in both directions. We therefore treat this experiment as an overlapping-label transfer diagnostic rather than broad proof of cross-domain generalization.

The iMiGUE$\rightarrow$SMG target contains only 22 samples, so one additional correct prediction changes Accuracy by 4.55 points. Qwen3-VL-8B + GMoT reaches 86.36\% Accuracy (19/22), compared with 81.82\% (18/22) for the vanilla backbone. The target is also highly skewed, with 15 Folding arms, 4 Playing or adjusting hair, 2 Touching or covering suprasternal notch, and 1 Touching or scratching neck samples. We therefore report Accuracy, Macro-F1, and Weighted-F1 together and avoid statistical-strength claims from this split. On SMG$\rightarrow$iMiGUE, the small Weighted-F1 fluctuation is concentrated in a class with only three test samples, further motivating per-class and class-support-aware interpretation.

Qwen2.5-VL-7B+\textbf{GMoT} remains a strong baseline. It raises iMiGUE$\rightarrow$SMG Accuracy to 77.27\% and Weighted-F1 to 0.7786, but it still trails Qwen3-VL-8B+\textbf{GMoT} on both metrics in that direction; on SMG$\rightarrow$iMiGUE it reaches 52.24\% Accuracy, and both Macro-F1 and Weighted-F1 remain lower. The comparison with Qwen2.5-VL-7B (w/o GMoT) further highlights the contribution of the motion token branch: removing GMoT reduces iMiGUE$\rightarrow$SMG Accuracy from 77.27\% to 50.00\% and Weighted-F1 from 0.7786 to 0.5578, while SMG$\rightarrow$iMiGUE Weighted-F1 drops from 0.5378 to 0.5034.

The task-specific MM-Gesture RGB baseline~\cite{gu2025mm} also drops sharply under the same direct-transfer splits, which shows that the diagnostic is challenging for a dedicated recognition model as well as general MLLMs. Within this restricted protocol, GMoT improves accuracy and Macro-F1 over its Qwen3-VL backbone, but the mixed Weighted-F1 results and small target support remain explicit limitations. Appendix~C provides the split audit and extended protocol.

Additional repeat-run stability and controlled corruption results are reported in Appendix~C.2.

\noindent\textbf{Analysis of Reasoning Ability.}
An anatomically grounded micro-gesture rationale should mention the body region relevant to the predicted class. We therefore define \textbf{Body-Region Grounding (BRG) Recall} by mapping each class $c$ to a region-specific keyword set $K_c$ and checking whether the generated reasoning trace $T_i$ overlaps with $K_c$:
\begin{equation}
    BRG = \frac{1}{N_{correct}} \sum_{i=1}^{N_{correct}} \mathbb{I}(K_c \cap T_i \neq \emptyset)
\end{equation}
where $\mathbb{I}$ is the indicator function. BRG is computed only over correctly classified samples, so it is a lexical anatomical-grounding proxy rather than a general measure of causal reasoning faithfulness. We report \textit{Avg. Length}, the average number of words inside the \texttt{<think>} block, alongside BRG to expose the potential verbosity confound.

\begin{table}[t]
\centering
\footnotesize
\setlength{\tabcolsep}{3pt}
\caption{Reasoning evaluation on iMiGUE~\cite{liu_imigue_2021} after policy refinement. BRG is the body-region keyword-overlap rate conditioned on a correct prediction; Avg. Len. is a diagnostic rather than an optimization target.}
\label{tab:reasoning_metrics}

\begin{tabular}{l c c c}
\toprule
Method & Acc. (\%) $\uparrow$ & BRG $\uparrow$ & Avg. Len. \\
\midrule
GPT-4o~\cite{hurst2024gpt} & 56.89 & 56.21 & 30.91 \\
\midrule
Qwen3-VL-8B~\cite{bai2025qwen3vltechnicalreport} & 62.96 & 96.82 & 45.08\\
Qwen3-VL-8B+GMoT & \textbf{67.32} & \textbf{97.70} & \textbf{64.84}\\
\bottomrule
\end{tabular}%

\end{table}
\Cref{tab:reasoning_metrics} shows that both Qwen3-VL variants attain high BRG among their correct predictions. GMoT improves accuracy from 62.96\% to 67.32\% after the same policy-refinement stage, while BRG changes from 96.82\% to 97.70\% and average rationale length increases from 45.08 to 64.84 words. Because a longer rationale has more opportunities to contain a body-region keyword, the length increase is reported as a confound and is not itself treated as evidence of better reasoning. A fully matched-length BRG comparison is not available from the current logs.

We also move the blind human check into the main evaluation discussion. On 181 accepted SMG descriptions, both annotators agreed that 137 were visually grounded. For severe hallucination, 44 samples were flagged by both annotators and one additional sample was flagged by one annotator, yielding 89 positive votes out of 362. Under a conservative item-level rule that treats the single disagreement as positive, the severe-hallucination count is 45/181 (24.86\%; 95\% Wilson CI: 19.13\%--31.64\%). This human check supports the use of the annotations as intermediate supervision but does not make them gold-standard rationales. Appendix~A gives the sampling and agreement details.

\subsection{Qualitative Analysis}

To better understand why \textbf{GMoT} improves micro-gesture reasoning, we qualitatively compare the generated rationales and predictions across representative MLLMs. \Cref{tab:qual_reasoning} shows a challenging clip labeled as \emph{Shake double shoulders}, where the discriminative motion is subtle and localized around the shoulder region, while the overall posture contains a strong yet misleading cue of folded arms. As shown in \Cref{tab:qual_reasoning}, general-purpose MLLMs often default to salient appearance cues and produce rationales that are temporally insensitive. Instead of isolating the crucial shoulder dynamics, they describe the clip as largely unchanged over time and consequently predict visually plausible but motion-agnostic classes such as \emph{Holding arms}, \emph{Touching arm}, or \emph{Crossing arms}. 

In contrast, our method produces a motion-grounded rationale that explicitly attributes the decision to shoulder dynamics, such as the raised and slightly hunched shoulders, and correctly predicts \emph{Shake double shoulders}. This is consistent with our design goal, which distills motion-relevant evidence into a compact token sequence before temporal modeling. By emphasizing regions that change over time and suppressing static background cues, GMoT reduces distraction from dominant but non-discriminative appearance signals, yielding more motion-aligned explanations in this qualitative case. Appendix~D provides additional examples and limitations.

\begin{figure}[t]
    \centering
    \includegraphics[width=0.9\linewidth]{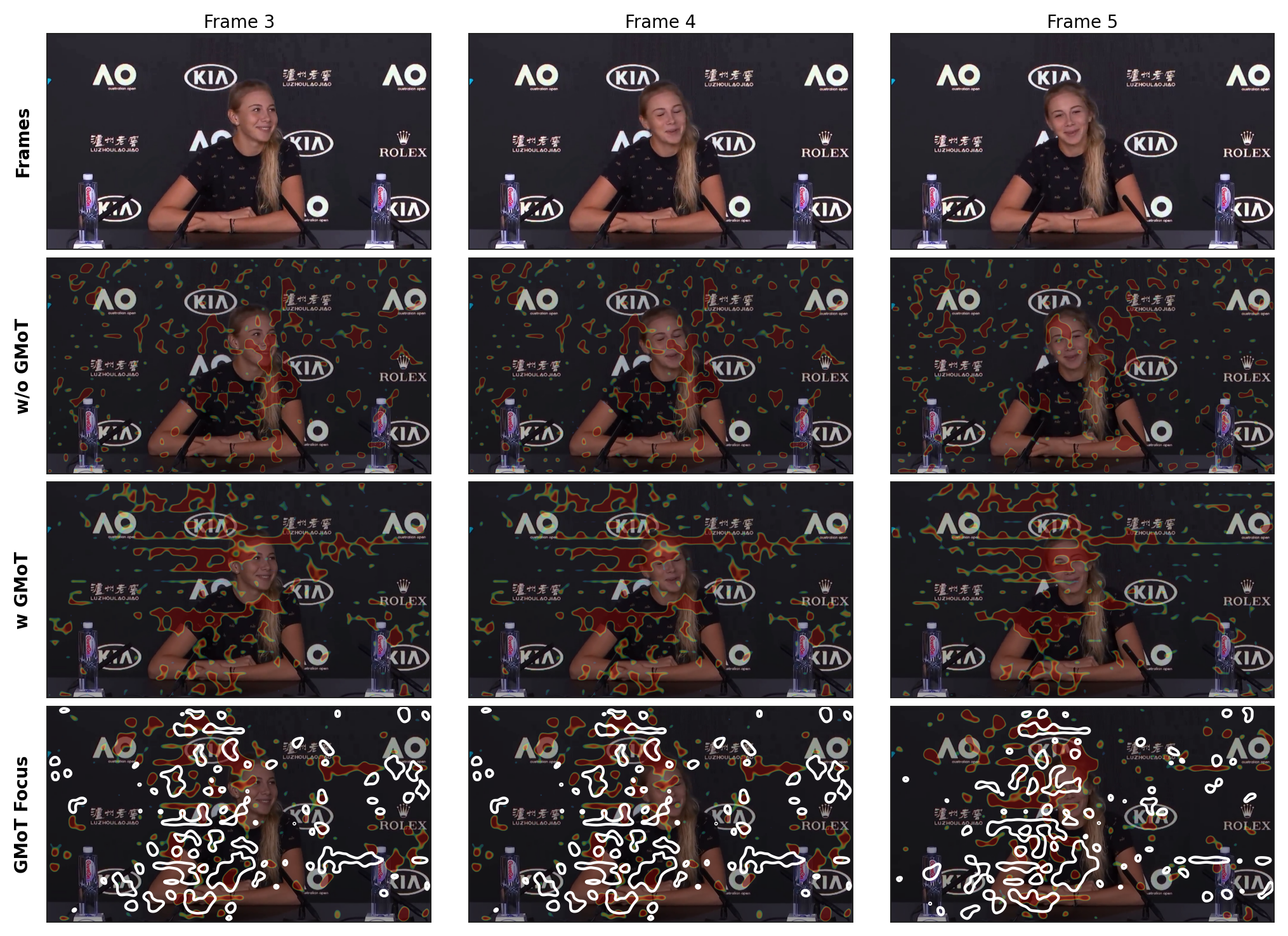}
    \caption{Qualitative visualization on a \textit{Moving torso} sample. Compared with w/o GMoT, the model with GMoT shows more concentrated visual responses on the head-torso region. GMoT Focus line denotes the response difference map between models.}
    \Description{Side-by-side qualitative response maps for a Moving torso clip. The GMoT response is more concentrated around the head and torso than the response without GMoT, and a difference map highlights the changed regions.}
    \label{fig:qualitative_comparison}
\end{figure}

\Cref{fig:qualitative_comparison} provides a qualitative comparison between models with and without GMoT on a representative `\textit{Moving torso}' sample. Compared with the model without GMoT, the model equipped with GMoT produces more concentrated visual responses around the upper-body motion region while reducing scattered activations on irrelevant background areas. The response difference map further indicates that GMoT mainly amplifies motion-relevant cues, suggesting its effectiveness in guiding the model toward subtle regions.

\begin{table*}[t]
\centering
\small
\caption{Qualitative reasoning comparison on a challenging \emph{Shake double shoulders} clip. Incorrect reasoning is highlighted in \bad{red} and the correct answer is highlighted in \good{green}.}
\label{tab:qual_reasoning}
\setlength{\tabcolsep}{5pt}
\renewcommand{\arraystretch}{1.15}

\begin{tabularx}{\textwidth}{L{0.19\textwidth}Y}
\toprule
\multirow{2}{0.19\textwidth}{\centering\textbf{Video\\Sample}}
& {%
   \setlength{\tabcolsep}{2pt}%
   \setlength{\imgw}{\dimexpr(\hsize-8\tabcolsep)/5\relax}%
   \begin{tabular}{@{}c c c c c@{}}
   \includegraphics[width=\imgw,height=1.35cm,keepaspectratio]{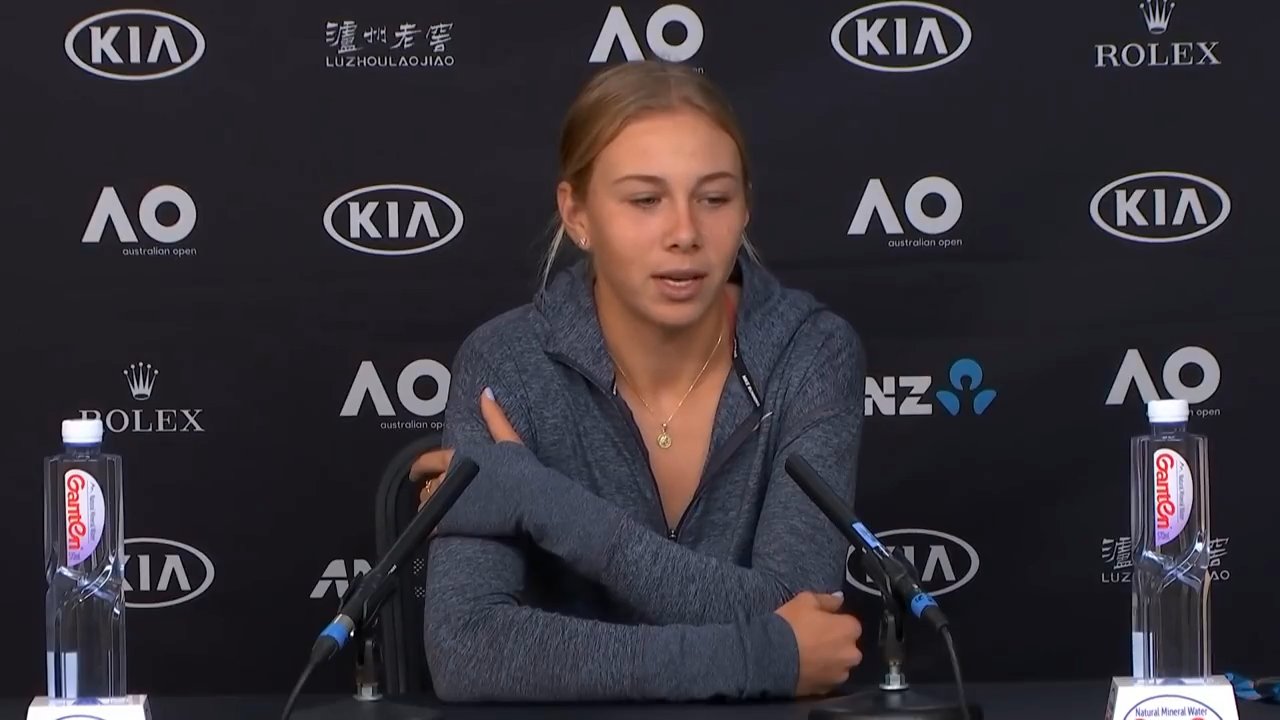} &
   \includegraphics[width=\imgw,height=1.35cm,keepaspectratio]{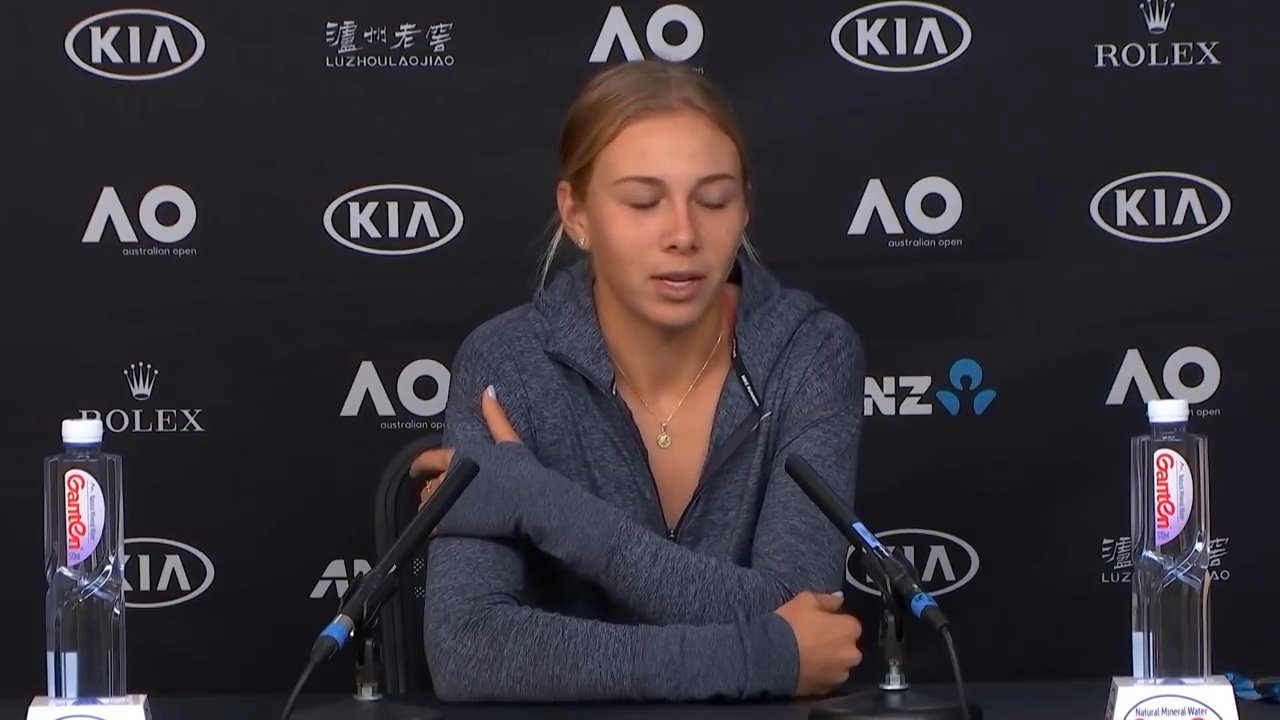} &
   \includegraphics[width=\imgw,height=1.35cm,keepaspectratio]{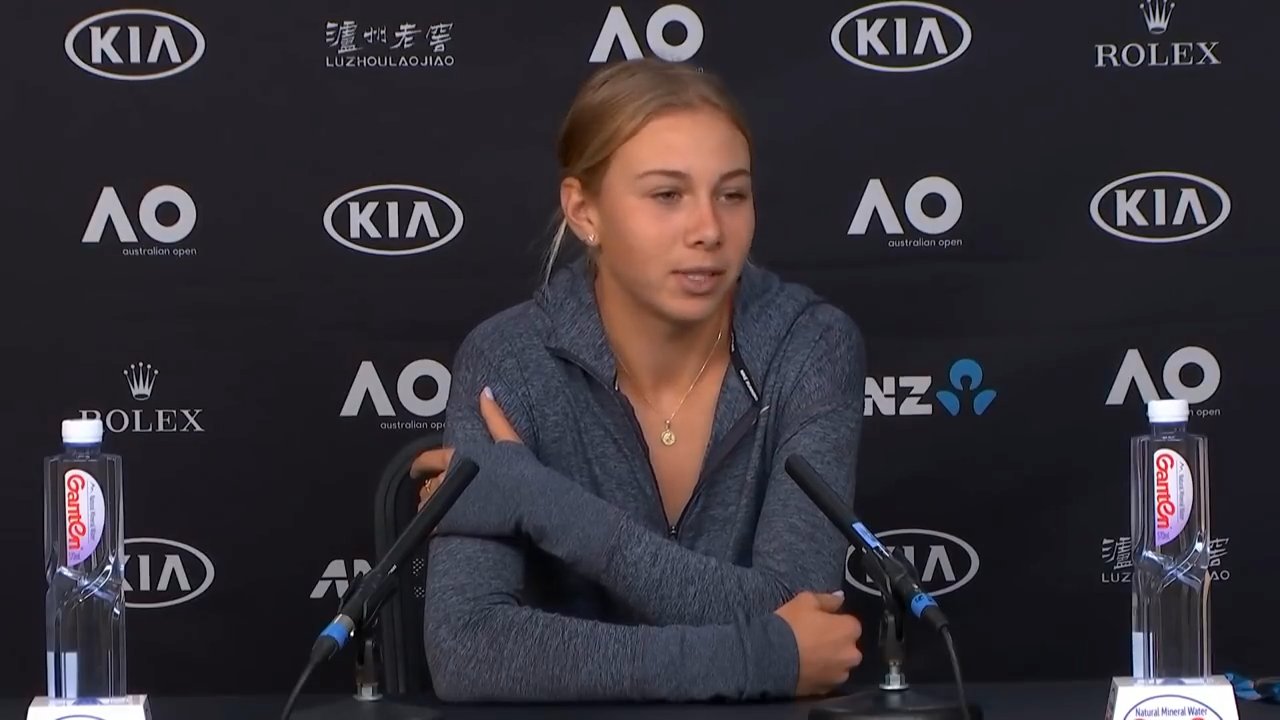} &
   \includegraphics[width=\imgw,height=1.35cm,keepaspectratio]{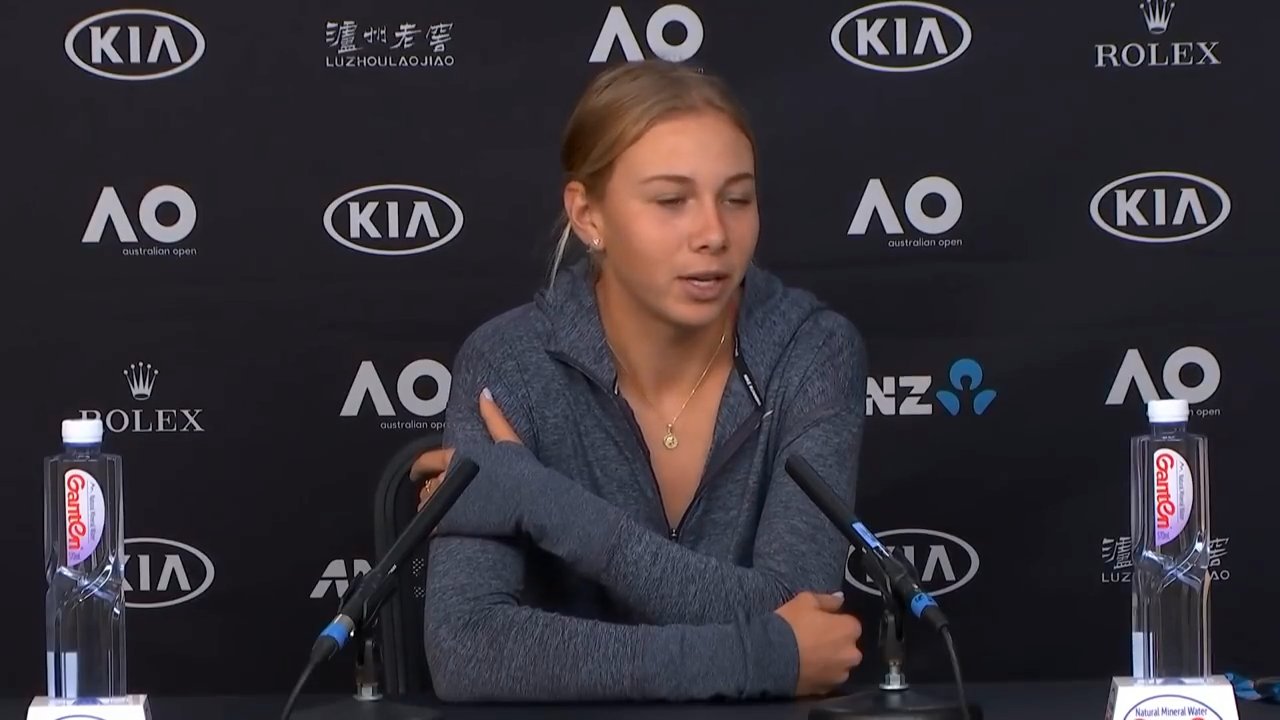} &
   \includegraphics[width=\imgw,height=1.35cm,keepaspectratio]{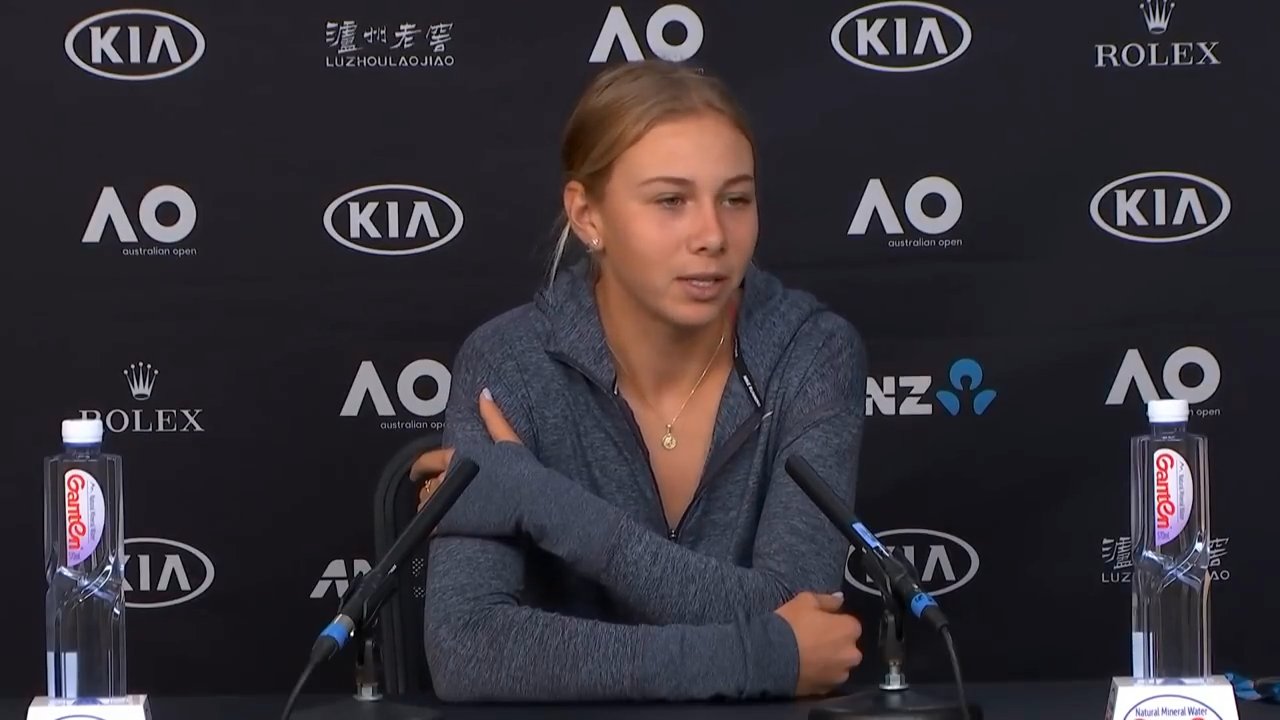}
   \end{tabular}
  }
\\
& Ground Truth: Shake double shoulders
\\ \midrule
\mname{ChatGPT}{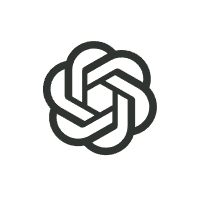}
& \tagt{<think>} Both arms are crossed in front of the torso, with one hand gripping the opposite upper arm/shoulder in a self-hug posture; no clear face/neck/leg-focused action. \tagt{</think>} \tagt{<answer>}\bad{ Holding arms}\tagt{</answer>}
\\ \midrule
\mname{Gemini}{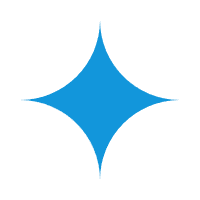}
& \tagt{<think>}\bad{I observe the person's right hand moving across her body to touch and rest upon her left upper arm.}\tagt{</think>}
\tagt{<answer>}\bad{Touching arm}\tagt{</answer>}
\\ \midrule
\mname{Qwen}{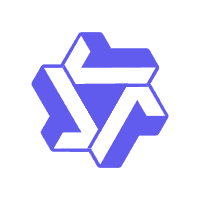}
& \tagt{<think>}The video shows a woman sitting at a press conference table. \bad{Her posture is static throughout the clip.} The most prominent feature is her arm position: her arms are folded across her chest. Her right hand is grasping her left upper arm, and her left arm is tucked underneath, holding her right arm. \tagt{</think>} \tagt{<answer>}\bad{Crossing arms}\tagt{</answer>}
\\ \midrule
\textbf{GMoT (Ours)}
& \tagt{<think>}\good{The person's shoulders are raised and slightly hunched, indicating a possible feeling of discomfort or unease.} The head is slightly tilted forward, and the upper body appears tense, with the arms crossed in front. This posture can be associated with a need to self-soothe or a moment of contemplation during the conversation. \tagt{</think>} \tagt{<answer>}\good{Shake double shoulders}\tagt{</answer>}
\\ \bottomrule
\end{tabularx}
\vspace{0.55em}

\noindent
\begin{minipage}[t]{0.38\textwidth}
\centering
\captionsetup{type=table}
\captionof{table}{Transfer study on Video-LLaMA2.}
\label{tab:videollama2_transfer}
\scriptsize
\setlength{\tabcolsep}{3pt}
\renewcommand{\arraystretch}{1.04}
\begin{tabularx}{\linewidth}{@{}>{\raggedright\arraybackslash}Xcc@{}}
\toprule
\textbf{Method} & \textbf{iMiGUE} & \textbf{SMG} \\
\midrule
VideoLLaMA2-Qwen2.5-7B & 39.74 & 51.80 \\
+ Motion Token & 39.52 & 50.33 \\
+ Motion Token + CoT mixed training & \textbf{42.75} & \textbf{52.13} \\
\bottomrule
\end{tabularx}
\end{minipage}\hfill
\begin{minipage}[t]{0.31\textwidth}
\centering
\captionsetup{type=table}
\captionof{table}{Stage-wise ablation on iMiGUE.}
\label{tab:stage_ablation}
\scriptsize
\setlength{\tabcolsep}{3pt}
\renewcommand{\arraystretch}{1.04}
\begin{tabularx}{\linewidth}{@{}>{\raggedright\arraybackslash}Xc@{}}
\toprule
Setting & \textbf{Acc. (\%)} \\
\midrule
S0: Qwen3-VL SFT w/o GMoT & 60.52 (0.00) \\
S1: S0 + Policy Refinement & 62.96 (\textcolor{red}{+2.44}) \\
\midrule
S2: GMoT warm-up + Video-Label Pairing & 59.85 (\textcolor{green}{-0.67}) \\
S3: S2 + CoT-format SFT & 61.19 (\textcolor{red}{+0.67}) \\
S4: S3 + Policy Refinement & 67.32 (\textcolor{red}{+6.80}) \\
\bottomrule
\end{tabularx}
\end{minipage}\hfill
\begin{minipage}[t]{0.26\textwidth}
\centering
\captionsetup{type=table}
\captionof{table}{GMoT component ablation on iMiGUE.}
\label{tab:gmot_ablation}
\scriptsize
\setlength{\tabcolsep}{3pt}
\renewcommand{\arraystretch}{1.04}
\begin{tabularx}{\linewidth}{@{}>{\raggedright\arraybackslash}Xc@{}}
\toprule
Setting & \textbf{Acc. (\%)} \\
\midrule
Full GMoT (Stage 2) & 61.19 \\
\midrule
w/o Spatial Scorer & 56.94 (\textcolor{red}{-4.25}) \\
w/o Temporal Diff & 55.65 (\textbf{\textcolor{red}{-5.54}}) \\
w/o Semantic Gate & 56.28 (\textcolor{red}{-4.91}) \\
\bottomrule
\end{tabularx}
\end{minipage}
\end{table*}
\subsection{Ablation Study}
In the ablation study, we ask whether the four-stage strategy is necessary and quantify the contribution of each GMoT component. The secondary Video-LLaMA2 transfer check is discussed in Appendix~C.
\subsubsection{Stage-wise ablation.}
\Cref{tab:stage_ablation} progressively enables each stage while keeping the backbone, sampling protocol, and optimization budget comparable. Video-label pairing alone slightly reduces accuracy to 59.85\%, CoT-format SFT recovers it to 61.19\%, and policy refinement then reaches 67.32\%. The larger gain from refinement after reasoning-aware initialization supports the staged training design. Appendix~C provides the extended interpretation.



\section{Limitations}
The iMiGUE$\rightarrow$SMG diagnostic has only 22 target samples and a highly skewed class distribution, so its result should not be generalized to broad cross-domain robustness. BRG is a lexical body-region proxy conditioned on correct predictions and can be affected by rationale length; it does not establish causal faithfulness. The sampled annotation audit is category-skewed and leaves a non-negligible residual grounding and hallucination error rate. The corruption study covers three controlled operators rather than unconstrained real-world degradation, and adjacent-frame differencing is not a complete model of long-range temporal dependency. Future work should evaluate larger aligned cross-domain sets, matched-length grounding metrics, and broader natural corruptions.

\section{Conclusion}
We propose GMoT, a motion-aware tokenization module that distills subtle kinematics into a compact sequence before language reasoning. Coupled with semi-automatic rationale supervision and reward-guided refinement, GMoT improves in-domain recognition on iMiGUE and SMG, retains gains under the tested corruptions, and improves Accuracy and Macro-F1 in a restricted overlapping-label transfer diagnostic. Its generated rationales maintain high lexical anatomical grounding, while the explicit limitations of BRG, annotation quality, and small-split transfer remain part of the reported evidence.

\begin{acks}
This work was supported by the National Natural Science Foundation of China (Grant No.~62576076), the CCF-Tencent Rhino-Bird Open Research Fund, and the Guangdong Research Team for Communication and Sensing Integrated with Intelligent Computing (Project No.~2024KCXTD047). Computational resources were provided by the SongShan Lake HPC Center (SSL-HPC) at Great Bay University.
\end{acks}

\bibliographystyle{ACM-Reference-Format}
\bibliography{main}

\end{document}